\documentclass{article}

\usepackage[utf8]{inputenc}
\usepackage{amsmath,amssymb}
\usepackage{graphicx}
\usepackage{booktabs}
\usepackage{float}
\usepackage{hyperref}
\usepackage{xcolor}
\usepackage{makecell}
\usepackage{multirow}
\usepackage{authblk}
\usepackage{amsthm}
\usepackage{mdframed}
\usepackage{mathtools}

\title{Formalising the Logit Shift Induced by LoRA: A Technical Note}

\author{
    Xiang Shi$^{1,3}$ \quad
    Shuaizhi Cheng$^{2,3}$ \quad
    Mingwei Li$^{3}$
}
\affil{
    $^1$ Imperial College London \\
    $^2$ Harbin Institute of Technology \\
    $^3$ KigLand Machine Learning Lab \\
    \texttt{\{kevin, shuaizhi, remi\}@kig.land}
}

\DeclareMathOperator{\tr}{tr}

\newtheorem{assumption}{Assumption}
\newtheorem{proposition}{Proposition}
\newtheorem{remark}{Remark}

\begin{document}
\setlength{\parindent}{0pt}
\newcommand{\trans}[0]{^\top}

\maketitle

\begin{abstract}
This technical note provides a first-order formalisation of the logit shift and fact-margin change induced by Low-Rank Adaptation (LoRA).
Using a first-order Fr\'echet approximation around the base model trajectory, we show that the multi-layer LoRA
effect can be decomposed into a linear summation of layerwise contributions and a higher-order remainder term
representing inter-layer coupling.
\end{abstract}

\section{Introduction}

Low-Rank Adaptation (LoRA) \cite{hu2022lora} is now a widely used approach to parameter-efficient fine-tuning (PEFT) for large Transformer models. Instead of updating the full parameter set, LoRA freezes the pretrained weights and inserts trainable low-rank matrices into selected linear modules, such as attention and MLP projections. This substantially reduces the number of trainable parameters and the memory cost of adaptation, while retaining strong empirical performance across a range of downstream settings.

Despite its practical success, the mechanism by which LoRA modifies a model's final predictions is still not fully understood. At the parameter level, LoRA introduces an additive low-rank perturbation to a weight matrix. However, in a deep Transformer, such a perturbation is propagated through a highly nonlinear computation graph, so its effect on the final logits is not simply additive or locally obvious. For this reason, it is useful to develop a principled approximation that connects a layerwise LoRA perturbation to its downstream effect on the model output.

This question is particularly relevant in settings where one seeks precise and interpretable control over model behaviour. For example, in knowledge editing, one would like to understand how a local parameter update changes the model's preference between a pretrained fact and a document-supported alternative. More broadly, in mechanistic interpretability and PEFT analysis, it is important to characterise which intermediate representations are affected by LoRA, along which directions they are perturbed, and how these perturbations are amplified or attenuated by the remaining layers.

In this note, we provide a local first-order analysis of the logit shift induced by LoRA. Using a Fr\'echet expansion around the base-model trajectory, we derive an explicit first-order expression for the logit change caused by a LoRA perturbation at a single layer. We then extend the analysis to the multi-layer setting, where the total first-order effect is shown to decompose into a sum of layerwise contributions, with inter-layer coupling captured by a higher-order remainder term. Finally, we apply the same framework to the fact margin between two candidate outputs, yielding a formal criterion for when a LoRA update is sufficient to reverse the model's preference.

Our goal is not to claim a globally exact decomposition of LoRA behaviour in nonlinear networks, but rather to provide a rigorous local characterisation of its leading-order effect on model logits. We hope this formulation can serve as a useful theoretical basis for analysing, diagnosing, and interpreting LoRA-based adaptation.
\section{Preliminary}

We consider a linear layer located at layer $l$ of the transformer,
\[
W_l : \mathbb{R}^{d_{\mathrm{in}}} \to \mathbb{R}^{d_{\mathrm{out}}},
\qquad
z_{l+1} = W_l z_l,
\]
where
$
z_l \in \mathbb{R}^{d_{\mathrm{in}}},
W_l \in \mathbb{R}^{d_{\mathrm{out}} \times d_{\mathrm{in}}},
z_{l+1} \in \mathbb{R}^{d_{\mathrm{out}}}.
$\\

Under LoRA \cite{hu2022lora}, the weight matrix of this layer is rewritten as
\[
\widetilde{W}_l = W_l + \Delta W_l,
\qquad
\Delta W_l = \frac{\alpha}{r} B_l A_l,
\]
where
$
B_l \in \mathbb{R}^{d_{\mathrm{out}} \times r},
A_l \in \mathbb{R}^{r \times d_{\mathrm{in}}}
$.
$\alpha \in \mathbb{R}$ denotes the LoRA scale, and $r \in \mathbb{N}^+$ denotes the LoRA rank.\\

Therefore, the increment in the output of this layer is
\[
\widetilde{z}_{l+1}
=
\widetilde{W}_l z_l
=
(W_l + \Delta W_l) z_l
=
z_{l+1} + \Delta W_l z_l,
\]
that is,
\[
\delta z_{l+1} := \widetilde{z}_{l+1} - z_{l+1} = \Delta W_l z_l.
\]

\begin{mdframed}[backgroundcolor=green!5]
\textbf{Notation:} In the following content, the notation $\tilde{\cdot}$ denotes the corresponding quantity after LoRA is applied.
\end{mdframed}

For an input $x$, let $h_L(x)$ denote the residual representation at the final layer. For a specific candidate token $y$, its logit is defined as
\[
\ell(y;x) = u_y^\top h_L(x),
\]
where $u_y$ is the unembedding vector corresponding to token $y$.

\begin{mdframed}[backgroundcolor=yellow!15]
\textbf{Note:} $h_L(x)$ is determined by the model state and does not itself depend on the candidate token $y$; the role of $y$ enters only through the readout direction $u_y$.
\end{mdframed}

\begin{mdframed}[
    backgroundcolor=blue!5,
    frametitle={First-Order Approximation},
    frametitlerule=true,
    frametitlebackgroundcolor=blue!10
]
\textbf{First-order expansion of a scalar-valued function:}
If $f:\mathbb{R}^n\to\mathbb{R}$ is differentiable at $u$, then for a small perturbation $v$,
\[
f(u+v)=f(u)+\nabla f(u)^\top v+o(\|v\|).
\]

\textbf{First-order expansion of a vector-valued function:}
If $F:\mathbb{R}^n\to\mathbb{R}^m$ is Fr\'echet differentiable at $u$, then there exists a linear map $DF(u):\mathbb{R}^n\to\mathbb{R}^m$ and a remainder term $r_F(v)$ such that
\[
F(u+v)=F(u)+DF(u)[v]+r_F(v),
\qquad
\frac{\|r_F(v)\|}{\|v\|}\to 0
\quad (v\to 0).
\]
In finite-dimensional Euclidean spaces, $DF(u)$ can be represented by the Jacobian matrix, and hence one may also write
\[
F(u+v)=F(u)+J_F(u)\,v+r_F(v),
\qquad
r_F(v)=o(\|v\|).
\]

\textbf{First-order expansion with a matrix-valued variable:}
If $g:\mathbb{R}^{m\times n}\to\mathbb{R}$ is differentiable at $M$, then for a small perturbation $N$,
\[
g(M+N)=g(M)+\langle \nabla g(M),N\rangle_F + o(\|N\|_F),
\]
where
$
\langle A,B\rangle_F := \tr(A^\top B)
$
is the Frobenius inner product.
\end{mdframed}

\section{First-Order Logit Shift Induced by LoRA}

We study the first-order effect of LoRA applied at layer $l$ on the final logit.\\

Define the map propagating the intermediate representation at layer $l+1$ to the final residual representation:
\[
F_l : \mathbb{R}^{d_{\mathrm{out}}} \to \mathbb{R}^{d_{\mathrm{model}}},
\qquad
h_L = F_l(z_{l+1}).
\]
After LoRA is applied, we have
\[
\tilde{h}_L
=
F_l(\widetilde{z}_{l+1})
=
F_l(z_{l+1}+\delta z_{l+1}).
\]

\begin{assumption}[Local differentiability at the base trajectory]
The map $F_l$ is Fr\'echet differentiable at the point $z_{l+1}$ along the base-model trajectory. Denote its derivative by
\[
DF_l(z_{l+1}) : \mathbb{R}^{d_{\mathrm{out}}} \to \mathbb{R}^{d_{\mathrm{model}}}.
\]
In coordinates, let the corresponding Jacobian be
\[
J_{l+1\to L}(x) := J_{F_l}(z_{l+1}).
\]
\end{assumption}

Under the above assumption, there exists a remainder function $r_l(v)$ such that
\[
F_l(z_{l+1}+v)-F_l(z_{l+1})
=
J_{l+1\to L}(x)\,v + r_l(v),
\qquad
\frac{\|r_l(v)\|}{\|v\|}\to 0
\quad (v\to 0).
\]
Substituting $v=\delta z_{l+1}=\Delta W_l z_l$, we obtain the exact expansion of the final residual representation:
\[
\delta h_L
:=
\tilde{h}_L-h_L
=
J_{l+1\to L}(x)\,\Delta W_l z_l + r_l(\Delta W_l z_l).
\]

Since the logit is
\[
\ell(y;x)=u_y^\top h_L,
\]
its variation satisfies
\begin{align*}
\delta \ell_y(x)
&:=
\tilde{\ell}(y;x)-\ell(y;x) \\
&=
 u_y^\top(\tilde{h}_L-h_L) \\
&=
 u_y^\top J_{l+1\to L}(x)\,\Delta W_l z_l + u_y^\top r_l(\Delta W_l z_l) \\
&=
 \frac{\alpha}{r}\,
 u_y^\top J_{l+1\to L}(x)\,B_lA_l z_l
 + u_y^\top r_l\!\left(\frac{\alpha}{r} B_lA_l z_l\right).
\end{align*}

\begin{proposition}[Single-layer first-order logit shift]
Under Assumption 1, as $\|\Delta W_l z_l\|\to 0$, one has
\[
\delta \ell_y(x)
=
\frac{\alpha}{r}
 u_y^\top J_{l+1\to L}(x) B_lA_l z_l
 + o\!\left(\|\Delta W_l z_l\|\right).
\]
Hence, the leading first-order term is
\[
\frac{\alpha}{r}
 u_y^\top J_{l+1\to L}(x) B_lA_l z_l.
\]
\end{proposition}

This shows that, in a first-order approximation around the forward trajectory of the base model, the effect of LoRA at layer $l$ on the logit of token $y$ is determined by three components:

\begin{enumerate}
    \item the local representation $z_l$ at the current layer;
    \item the LoRA injection direction $B_lA_l z_l$;
    \item the sensitivity of the downstream network in propagating this perturbation to the final readout, namely $J_{l+1\to L}(x)$.
\end{enumerate}

\section{Multiple LoRA Layers}

We now consider a collection of layers $S$ to which LoRA is applied. Let the perturbation at each layer be
\[
\Delta W_l = \frac{\alpha_l}{r_l} B_lA_l,
\qquad l\in S.
\]

To state the multi-layer sum rigorously, we collect the perturbations across all layers into a joint variable
\[
\Delta := (\Delta W_l)_{l\in S}.
\]
Let $G_y(\Delta)$ denote the final logit of token $y$ after these perturbations are applied simultaneously. The base model then corresponds to $\Delta=0$.

\begin{assumption}[Joint differentiability with respect to all LoRA perturbations]
The map $G_y$ is Fr\'echet differentiable at $\Delta=0$.
\end{assumption}

By Fr\'echet differentiability, there exist a linear map $DG_y(0)$ and a remainder term $R_y(\Delta)$ such that
\[
G_y(\Delta)-G_y(0)=DG_y(0)[\Delta]+R_y(\Delta),
\qquad
\frac{|R_y(\Delta)|}{\|\Delta\|}\to 0
\quad (\Delta\to 0).
\]
Since $DG_y(0)$ is linear, its action on the joint perturbation can be written as the sum of the first-order contributions along each coordinate direction:
\[
DG_y(0)[\Delta]
=
\sum_{l\in S} DG_y(0)[\iota_l(\Delta W_l)],
\]
where $\iota_l$ denotes the embedding map that is nonzero only in the $l$-th coordinate.

If the first-order derivative along each coordinate direction is identified with the leading term in the single-layer formula, then we obtain
\[
\tilde{\ell}(y;x)-\ell(y;x)
=
\sum_{l\in S}
 u_y^\top J_{l+1\to L}(x)\,\Delta W_l z_l
 + R_y(\Delta),
\qquad
R_y(\Delta)=o(\|\Delta\|).
\]
Expanding further yields
\[
\tilde{\ell}(y;x)-\ell(y;x)
=
\sum_{l\in S}
 \frac{\alpha_l}{r_l}
 u_y^\top J_{l+1\to L}(x)\,B_lA_l z_l
 + R_y(\Delta).
\]

If all layers share the same LoRA scale $\alpha$ and rank $r$, this can be written more compactly as
\[
\tilde{\ell}(y;x)-\ell(y;x)
=
\frac{\alpha}{r}
\sum_{l\in S}
 u_y^\top J_{l+1\to L}(x)\,B_lA_l z_l
 + R_y(\Delta),
\qquad
R_y(\Delta)=o(\|\Delta\|).
\]

\begin{remark}
The summation over layers here does not mean that the final logit is exactly decomposed into a sum of mutually independent layerwise logits. Rather, it means that, when one performs a first-order expansion around the base-model parameters with respect to the joint perturbation, the first-order marginal contributions of different layers add up by linearity of the derivative, whereas inter-layer coupling effects are absorbed into the higher-order remainder term $R_y(\Delta)$.
\end{remark}

\section{Fact Margin}

We further consider the margin between the document-supported fact $y_{\mathrm{doc}}$ and the pretrained fact $y_{\mathrm{pre}}$. Define
\[
m(x)
:=
\tilde{\ell}(y_{\mathrm{doc}};x)
-
\tilde{\ell}(y_{\mathrm{pre}};x).
\]

We also define the original margin under the base model as
\[
m_0(x)
:=
\ell(y_{\mathrm{doc}};x)
-
\ell(y_{\mathrm{pre}};x).
\]

Applying the joint first-order expansion from the previous section separately to the two tokens and subtracting, we obtain
\begin{align*}
m(x)
&=
m_0(x)
+
\sum_{l\in S}
\Big(
 u_{\mathrm{doc}} - u_{\mathrm{pre}}
\Big)^\top
J_{l+1\to L}(x)\,\Delta W_l z_l
+
R_m(\Delta) \\
&=
m_0(x)
+
\sum_{l\in S}
\frac{\alpha_l}{r_l}
\Big(
 u_{\mathrm{doc}} - u_{\mathrm{pre}}
\Big)^\top
J_{l+1\to L}(x)\,B_lA_l z_l
+
R_m(\Delta),
\end{align*}
where
\[
R_m(\Delta)=o(\|\Delta\|)
\qquad (\Delta\to 0).
\]

If all layers share the same $\alpha$ and $r$, then this becomes
\[
m(x)
=
m_0(x)
+
\frac{\alpha}{r}
\sum_{l\in S}
\Big(
 u_{\mathrm{doc}} - u_{\mathrm{pre}}
\Big)^\top
J_{l+1\to L}(x)\,B_lA_l z_l
+
R_m(\Delta).
\]

Therefore, the margin consists of three components:

\begin{enumerate}
    \item the base model's original preference between the two candidate facts, namely $m_0(x)$;
    \item the first-order correction induced by LoRA after propagation through each layer, measured along the readout direction $u_{\mathrm{doc}} - u_{\mathrm{pre}}$;
    \item a higher-order remainder term satisfying $R_m(\Delta)=o(\|\Delta\|)$.
\end{enumerate}

In particular, as long as the first-order correction is large enough to overcome both the original negative margin of the base model and the higher-order remainder, i.e.,
\[
\frac{\alpha}{r}
\sum_{l\in S}
\Big(
 u_{\mathrm{doc}} - u_{\mathrm{pre}}
\Big)^\top
J_{l+1\to L}(x)\,B_lA_l z_l
>
-m_0(x)-R_m(\Delta),
\]
then $m(x)>0$, meaning that on this input the model prefers the document-supported fact over the pretrained fact.

\section{Remark on Scope of the Approximation}

The above derivation is, in essence, a local first-order approximation around the forward trajectory of the base model. Accordingly, its validity depends on the following conditions:

\begin{enumerate}
    \item in the single-layer case, $F_l$ is Fr\'echet differentiable at the base point $z_{l+1}$;
    \item in the multi-layer case, the total logit is Fr\'echet differentiable at $\Delta=0$ with respect to the joint perturbation variable $\Delta$;
    \item the perturbation is sufficiently small so that the higher-order remainder is negligible relative to the first-order term;
    \item cross-layer couplings are generally nonzero, but they are contained in the higher-order term $o(\|\Delta\|)$ rather than vanishing altogether.
\end{enumerate}

Therefore, this derivation should be understood as a \emph{local linear interpretation} of the mechanism by which LoRA influences logits: it rigorously characterises the leading first-order term together with the structure of the remainder, but it does not claim that the true nonlinear network can be globally decomposed with exactness under perturbations of arbitrary magnitude.

\bibliographystyle{plain}
\bibliography{MLPMathProofRef}

\end{document}